\documentclass[journal]{IEEEtran}
\usepackage[cmex10]{amsmath}
\usepackage{amsfonts}
\usepackage{graphicx}
\usepackage{array}
\usepackage{multirow}
\usepackage{epstopdf}

\usepackage{bm}
\usepackage{color}

\begin{document}
\title{Edge-guided Non-local Fully Convolutional Network for Salient Object Detection
}


\author{Zhengzheng~Tu, Yan~Ma, Chenglong~Li, Jin Tang, Bin Luo
 \thanks{Z. Tu, Y. Ma, C. Li, J. Tang, and B. Luo are with Key Lab of Intelligent Computing and Signal Processing of Ministry of Education, School of Computer Science and Technology, Anhui University, Hefei 230601, China, Email: zhengzhengahu@163.com, m17856174397@163.com, lcl1314@foxmail.com, tangjin@ahu.edu.cn. C. Li is also with Institute of Physical Science and Information Technology, Anhui University, Hefei 230601, China. (\emph{Corresponding author is Chenglong Li})}
\thanks{This research is jointly supported by the National Natural Science Foundation of China (No. 61602006, 61702002, 61872005, 61860206004), Natural Science Foundation of Anhui Province (1808085QF187), Open fund for Discipline Construction, Institute of Physical Science and Information Technology, Anhui University.
}}

\markboth{IEEE Transactions on xxx}%
{Shell \MakeLowercase{\textit{et al.}}: Bare Demo of IEEEtran.cls for IEEE Journals}
\maketitle

\begin{abstract}
Fully Convolutional Neural Network (FCN) has been widely applied to salient object detection recently by virtue of high-level semantic feature extraction,
but existing FCN-based methods still suffer from continuous striding and pooling operations leading to loss of spatial structure and blurred edges.
To maintain the clear edge structure of salient objects, we propose a novel Edge-guided Non-local FCN (ENFNet) to perform edge-guided feature learning for accurate salient object detection.
In a specific, we extract hierarchical global and local information in FCN to incorporate non-local features for effective feature representations.
To preserve good boundaries of salient objects, we propose a guidance block to embed edge prior knowledge into hierarchical feature maps.
The guidance block not only performs feature-wise manipulation but also spatial-wise transformation for effective edge embeddings.
Our model is trained on the MSRA-B dataset and tested on five popular benchmark datasets.
Comparing with the state-of-the-art methods, the proposed method achieves the best performance on all datasets.
\end{abstract}


\section{Introduction}
\label{sec::introduction}
\IEEEPARstart{S}{aliency} detection is generally  divided into two categories: salient object detection or eye fixation prediction, with the purpose of extracting the most predominant objects or predicting human eye attended locations corresponding to informative regions in an image.
With the rapid development of deep learning, saliency detection has made great progress in recent years.
As a preprocessing step, saliency detection is helpful for many applications in the computer vision tasks, such as image segmentation~\cite{Donoser2010Saliency,Li16sold,li2017weighted}, scene classification~\cite{Ren2014Region}, visual tracking ~\cite{borji2012adaptive,li2016learning,Li18pami}, and person re-identification~\cite{zhao2013unsupervised}.

\begin{figure}[htbp]
\centering
\includegraphics[width=3.5in]{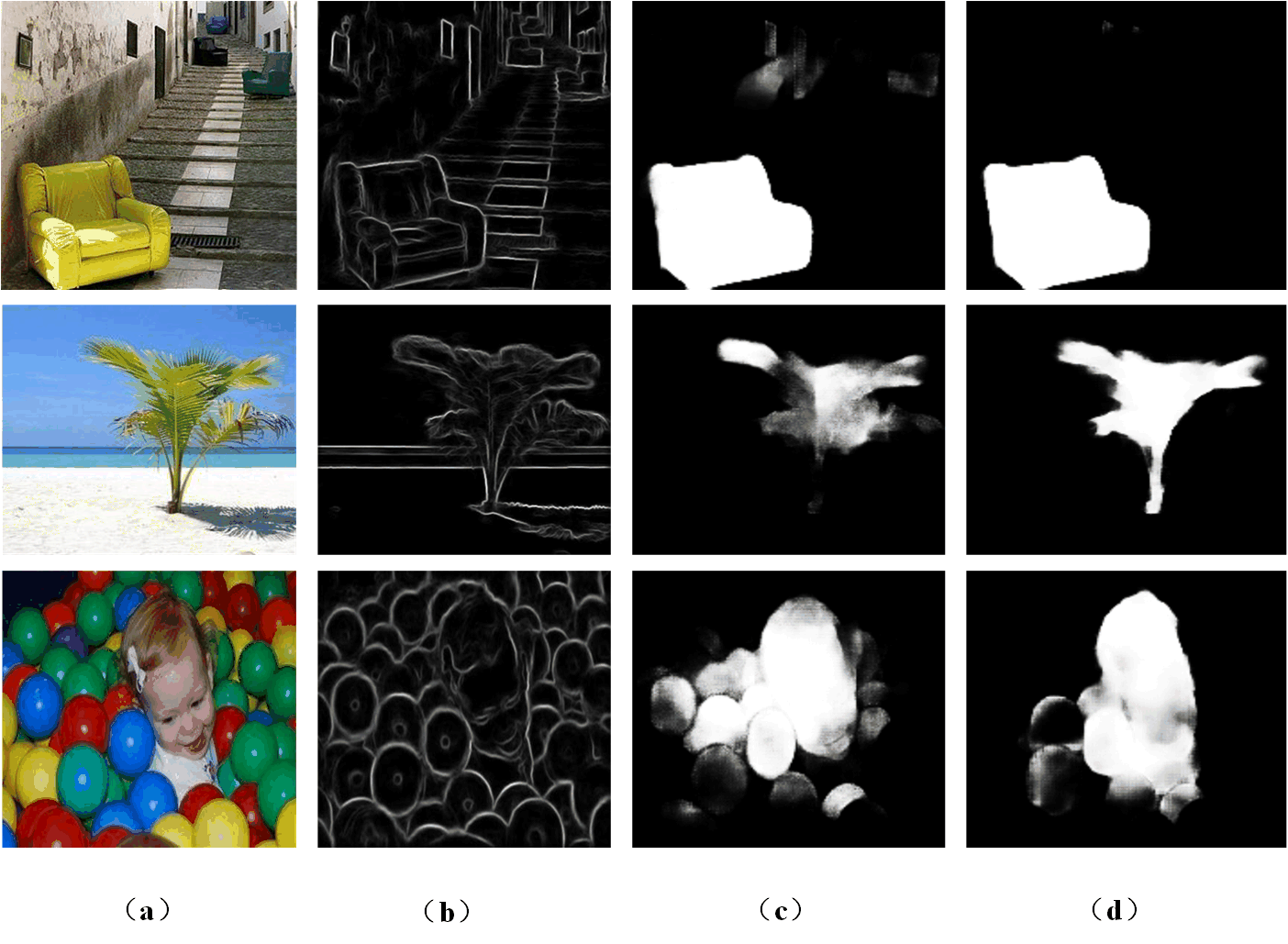}
\caption{Illustration of the importance of edge information to saliency detection. 
(a) Input images. 
(b) Edge maps. 
(c) Saliency results without edge embeddings. 
(d) Saliency results with edge embeddings.}
\label{1}
\end{figure}

The saliency of each region is usually defined as the weight of the distance between the region and other regions of the image in most traditional methods, and often based on hand-crafted features and some priors ~\cite{perazzi2012saliency, yang2013saliency, yan2013hierarchical}. 
%
%
In recent years, with the success of Fully Convolutional Network (FCN) in the field of computer vision, deep learning methods has become a promising alternative to salient object detection. 
Existing methods based on the FCN~\cite{LeeTK16,lee2016deep,WangLRY15,Li2016Deep,ShenW12,HouCHBTT17,Zhang2017Amulet,ChenZ0L17,Luo2017Non} have shown good performance, since the features obtained at deeper layers typically contain stronger semantic information and the global context-aware information, which are beneficial to locate salient regions even in complex scenes.
FCN promotes saliency detection by applying multiple convolution layers and pooling layers to increase the receptive field and automatically extract high-level semantic information, which plays an import role in this field. 
These continuous convolution layers and pooling operations gain large receptive fields and high representation ability but reduce the size of feature maps and lose the spatial structure, and deteriorate the edges of the salient objects. 
This is useful for some high-level tasks like classification and recognition, but unfortunately, it reduces the accuracy of low-level tasks which usually requires precise pixel activation, such as salient object boundaries.  

Some recent works notice this problem and propose some edge-aware saliency detection methods. 
For example, Zhang \emph{et al}.~\cite{Zhang2017Amulet} embed edge-aware feature maps achieved by shadow layers into the deep learning framework. 
Mukherjee \emph{et al}.~\cite{mukherjee2017salprop} construct a Bayesian probabilistic edge mapping which uses low-order edge features to assign a saliency value to the edge. 
However, boundary preserving still hasn't been solved well as lack of specific guidance for the boundary area.
As shown in Fig.~\ref{1}(c), the predictions around object boundaries are inaccurate as the essential fine details are lost caused by repeated strides and pooling operations. 
In a word, existing FCN-based methods still have the drawback that they are difficult to maintain spatial structure like clear boundaries.

To handle this problem, we propose a novel method based on an edge-guided non-local FCN, named ENFNet, for salient object detection.
We take the FCN model in~\cite{Luo2017Non} as the baseline of our ENFNet, 
and design an edge guidance block to perform feature-wise manipulation and spatial-wise transformation for effective edge embeddings.
In a specific, given the input image, we first extract the edge maps using a EdgeNet~\cite{Zitnick2014Edge}, and then employ the baseline FCN to perform salient object detection.
To embed the edge prior knowledge into the baseline FCN, we propose to use the spatially feature-wise affine transform algorithm~\cite{Wang2018Recovering} to guide feature learning.
An example of edge-guided saliency map generation by our method is shown in Fig.~\ref{1}.

In conventional models, the local contrast is widely used and plays an important role as salient objects should be different between foreground
and background areas.
While in deep models, the global context is very useful to model salient objects in the full image and complementary to the local contrast information.
To fully utilize both local contrast and global context, we design a hierarchical edge-guided non-local structure in our network.
In particular, features of each convolutional layer in FCN are first transformed into edge-guided features by edge prior knowledge, and then we employ the edge-guided features to generate local contrast features.
The global context features are computed by FCN and combined with the edge-aware local contrast features to produce the final saliency map.
We compare the proposed approach with many start-of-the-art saliency detection methods on the five benchmark datasets 
including HKU-IS~\cite{li2015visual}, PASCAL-S~\cite{li2014secrets}, DUT-OMRON~\cite{yang2013saliency}, ECSSD~\cite{yan2013hierarchical} and SOD~\cite{martin2001database}, and our method achieves the best performance under all evaluation metrics.

In short, the major contributions of this work are summarized as follow:\\

\begin{itemize}

\item We propose a novel edge-guided non-local fully convolutional network for salient object detection. The proposed network is able to embed the detailed edge information in a hierarchical manner and thus generate high-quality boundary-aware saliency maps.

\item We design an edge guidance block to incorporate the edge prior knowledge in the non-local feature learning framework. The designed block is simple yet effective and thus yields a state-of-the-art performance with a little impact on execution time.

\item Extensive experiments have proved that the proposed network is effective in producing good saliency maps with clear boundaries, and outperforms all compared state-of-the-art methods in all metrics.

\end{itemize}

\section{Related Work}
We will review two aspects of salient object detection approaches that are based on deep learning techniques and edge prior information.

\subsection{Deep learning based methods} 
Salient object detection can be regarded as a pixel-wise classification problem.
Although traditional salient object detection methods have their superiority, including no need training and simplicity, their overall performances are not as good as most deep learning methods. 
Deep learning based methods have achieved great improvement in salient object detection as they combine local and deep features and can be trained end-to-end.
In the state-of-the-art models of convolutional neural networl (CNN), feature selection between salient and non-salient regions is automatically accomplished by gradient descent algorithm. 
These models are made of a series of convolution layers and pooling until a softmax layer, which predicts the probability of each pixel belonging to objects. 
Recently, these methods make a great process by deep learning methods~\cite{li2015visual,WangWLZR16,KuenWW16,LeeTK16,lee2016deep,WangLRY15,Li2016Deep,ShenW12,HouCHBTT17,Zhang2017Amulet,ChenZ0L17}, which can be divided into region-based and Fully Convolutional Networks (FCN)-based methods.

{\flushleft \bf Region-based methods.}
The region-based deep learning methods use image patches as the basic processing units for saliency prediction. 
For example, Li et al.~\cite{li2015visual} utilize multi-scale features to capture contextual information and a classifier network to infer the saliency of each image segment.
Zhao et al.~\cite{Rui2015Saliency} propose a multi-context deep CNN framework benefiting from the global context of salient objects, as global context is conducive to modeling saliency of the image, while local context is conducive to estimating saliency of the region with rich features. 
Li et al.\cite{Li2016Deep} propose a deep contract network to combine a piece-wise stream and a pixel-level stream for saliency detection. 
And Wang et al.~\cite{Wang2015Deep} propose to train two deep neural networks to integrate global search and local estimation for salient object detection. 
The work in ~\cite{LeeTK16} uses a two stream framework to separately extract high-level features from VGG-net and low-level features such as color histogram. 
The fully-connected layers in CNNs is always to evaluate the saliency of every region~\cite{wang2018detect}. 
The region-based methods improve the performance over the ones based on hand-crafted features to a great extent, whereas they ignore the important spatial information.  
In addition, the region-based methods are time-consuming, as the network has to run many times.

\begin{figure*}[htbp]
\centering
\includegraphics[width=7.2in]{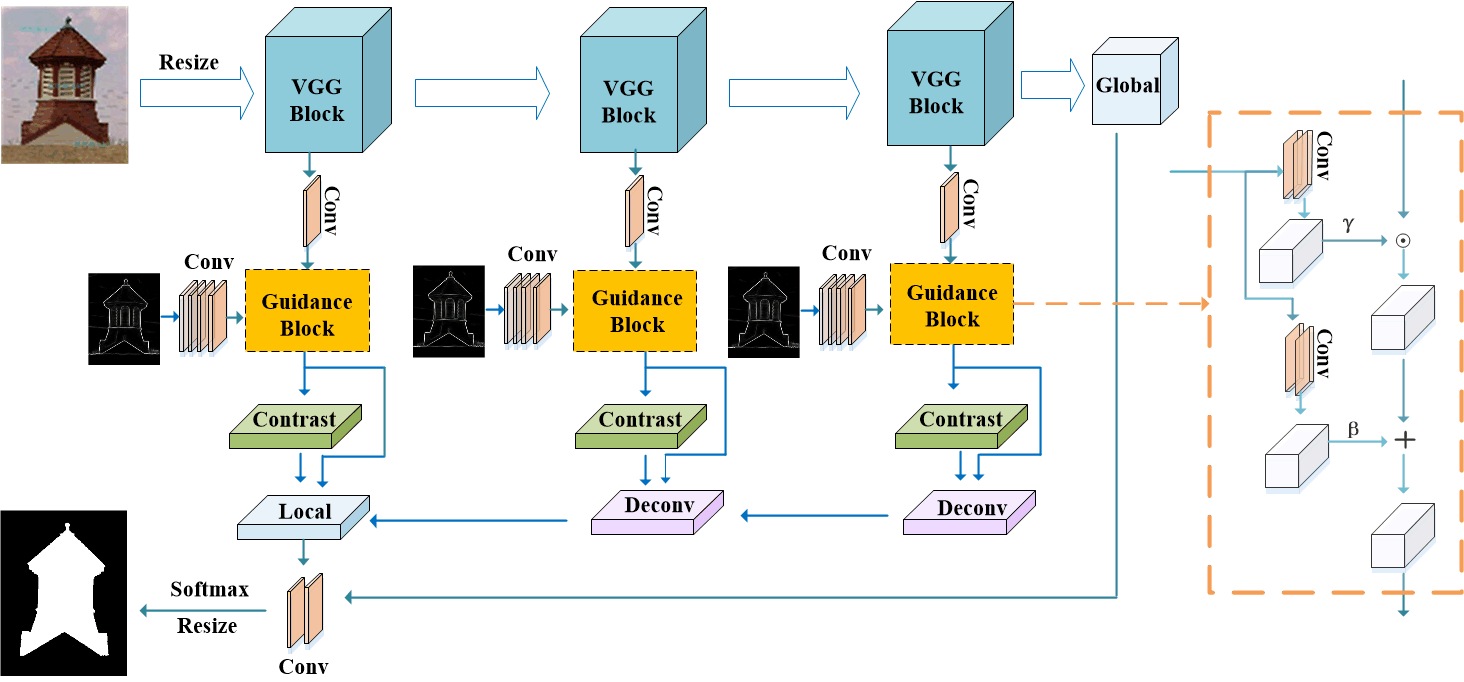}
\caption{Overall architecture of our proposed model. 
Our architecture is based on VGG-16~\cite{simonyan2014very} for better comparison with previous CNN-based methods. 
Herein, only three convolution blocks are listed for clarity. 
The dotted-line part is an edge guidance block, where the edge features are regarded as a condition information on feature learning.}
\label{2}
\end{figure*}

{\flushleft \bf FCN-based methods.}
To improve the efficiency of CNN methods, people utilize FCN to generate a pixel-wise prediction. 
The FCN-based methods abandon the fully connected layer in CNN. 
They can increase the computational efficiency, but lost the spatial information. 
For example, Li et al.~\cite{Li2016Deep} make use of segment-level spatial pooling stream and pixel-level fully convolutional stream to generate saliency estimation. 
The work in~\cite{WangWLZR16} designs a deep recurrent FCN to incorporate the coarse estimations as saliency priors and refines the generated saliency stage by stage. 
In HED~\cite{Xie2015Holistically}, the authors build skip-connections to employ multi-scale deep features for edge detection. 
The edge detection task is easier than saliency detection since it does not rely on much high-level semantic information. 
Thus, directly using skip-connections in salient object detection is unsatisfactory. 
Then, Hou et al.~\cite{Hou2016Deeply} propose dense short connections to skip-layers within the HED architecture to obtain rich multi-scale features for salient object detection. 
Another work by SRM~\cite{Wang2017A} proposes a pyramid pooling model and a stage-wise refinement model to integrate both global and local context information. 
Zhang at el.~\cite{Zhang2017Amulet} aggregate high level and low level features by concatenating feature maps directly, these multi-level features are beneficial to recover local details and locate salient objects. 
In addition, many methods attempt to find a way to fuse multi-scale features for better distinguishing salient and non-salient regions. 
However, without edge information, they can roughly detect the targets but cannot uniformly highlight the entire objects, and the estimated saliency maps suffer from the blur boundary.


\subsection{Exploiting Edge Information} 
To solve above-mentioned blurred boundary problem, some methods attempt to take advantage of edge information for generating saliency maps with clear boundaries. 
Mukherjee et al.~\cite{mukherjee2017salprop} construct a Bayesian probabilistic edge mapping and assign a salient value to the edge by using low-order edge features, then they learn a conditional random field to effectively combine these features with object/non-object labels for edge classification. 
The work in~\cite{Xiang2016Salient} and~\cite{wang2017edge} use edge information for detecting salient regions. 
For example, in ~\cite{Xiang2016Salient}, they exploit a pre-trained edge detection model to detect the edge of objects. 
Then, based on the detected edge, they segment the image into regions, and generate saliency map of every region through a mask-based Fast R-CNN~\cite{sappa2006edge}. 
The model in~\cite{zhang2017deep} uses three different labels including salient objects, salient object's boundaries and background. 
And in~\cite{Jiangjiang2019}, they add a edge detection branch into the pooling network, through joint training saliency model and edge detection, the details of salient objects are further sharpened. 
To emphasis the accuracy of saliency boundary detection, the model~\cite{zhang2017deep} takes the extra hand-craft edge feature as a complementary to preserve edge information effectively. 
Guan et. al~\cite{guan2019edge} propose an edge detection stream to combine multiple side outputs together through concatenation and uses a fusion layer that a $1\times1$ convolution to get the unified output. 
With the edge information added, these above models achieve good performance in preserving the boundary of salient object.

These methods usually use edge losses to preserve the edges of salient objects or fuse shallow features and edge features for simple feature aggregation. 
To make better use of edge information, from a global and local perspective, we not only add edge features in shallow layers but in deep layers. 
At the same time, our edge guidance block which includes a series of operations on edge features can be implemented feature-wise affine transform, which could more effectively retain spatial structure information.
Through using parallel edge features to guide saliency detection, we could obtain more accurate saliency maps with clearer object boundaries.

\section{Edge-guided Non-local FCN}

In this section, we will introduce the architecture structure of our proposed Edge-guided Non-local Fully convolutional Network (ENFNet). The overall structure of ENFNet is shown in Fig.~\ref{2}.

\subsection{Overview of Network Architecture}
As discussed above, we combine global context information and local structure information with different resolution details, and integrate edge priors into different resolution features to better locate the boundary of the salient object. 
Existing methods usually use complex backbone networks~\cite{deng2018r3net} to learn powerful features. 
To improve the efficiency, we adopt the VGG-16 network~\cite{simonyan2014very} as our backbone network like~\cite{Luo2017Non}. 
The image size of the input model is fixed to 352$\times$352, and the output saliency map is 176$\times$176. 
To obtain the saliency map of the same size as the input image, we use the bilinear interpolation method to upsample feature maps. 
However, when the foreground and background have the same contrast, and the object has a complex background, the predicted salient results are often not so good, such as unclear boundaries. 
Therefore, to preserve good boundaries of salient objects, we design an edge guidance block to embed edge prior knowledge into hierarchical feature maps for effective feature representations. 
The edge prior knowledge is beneficial to guide the results of saliency detection to possess good object boundaries as shown in Fig.~\ref{1}.

As shown in Fig.~\ref{2}, our ENFNet does not use the fully-connected layers because the task of saliency detection is focused on pixel-level prediction. 
We select the first five blocks of VGG-16 as side outputs, where each side output is followed by a convolution block, which is designed to obtain hierarchical multi-scale features ${\{X_{1}, X_{2},  X_{3}, X_{4}, X_{5} \}}$. 
Each convolution block contains two or three convolutional layers and a max pooling operation. 
The last block behind the backbone network contains three convolutional layers to computes global features $X_{G}$ that is the global context of the image.

The middle layers is we proposed edge guidance block, which is located on the right side of Fig.~\ref{2}. 
Details of the edge guidance block as shown in Table~\ref{tab1}. 
First, we extract edge maps using the existing method~\cite{Zitnick2014Edge}, and these edge maps $X^E_{i}$ are used as input of the edge guidance block. 
Second, the edge features $X^E_{i}$ and the hierarchical multi-scale features $X_{i}(i\in\{1, 2, 3, 4, 5\})$ are fused by the edge guidance block to generate edge-aware features $X^F_{i}$. 
Finally, the edge features $X^F_{i}$ are through average pooling operation to obtain the contrast features $X_{i}^{C}$. 
Because the salient objects are related to global or local contrast, the contrast features capture the difference between each region and its neighbors.
The last line of our model is a series of deconvolution blocks, and these deconvolution operation update the features maps from $11\times11$ to $176\times176$, for the convenience of concatenating feature maps $(X^F_{i}, X^C_{i})$ with different scales. 
The local blocks contain one convolution layer to compute the local feature $X^{L}$. 
In the final line we use two convolution layers and a softmax operation to compute the scores that equal to the predicted saliency map by fusing the local and global features.

\subsection{Hierarchical Non-local Structure}
Some works show that lower layers in convolutional neural networks can capture rich spatial information, while higher layers encode object-level information but are invariant to factors such as appearance and pose~\cite{pinheiro2016learning}. 
Generally speaking, the receptive field of lower layers is limited, it is unreasonable to require the network to perform dense prediction in the early stage. 
Performing feature extraction instead of feature classification at earlier stages, those extracted low-level features can provide spatial information for our final predictions at the bottom layer. 
Therefore, the multi-scale features are quite important to improve the performance of saliency detection. 
As shown in Fig.~\ref{2}, these convolution blocks below in side outputs are processed by the first five convolutional blocks of VGG-16. 
The function of these convolution blocks is to obtain different scale features ${\{X_{1}, X_{2}, X_{3}, X_{4}, X_{5}\}}$. 
Each of convolution block uses the kernel size of $3\times3$ and outputs features with 128 channels.

\begin{center}
\begin{table}
\caption{Details of the proposed Edge Guidance Block (EG Block).}\label{tab1}
\setlength{\tabcolsep}{2.0mm}{
\begin{tabular}{|c|c|c|c|c|c|}
\hline
Block & Layer & kernal & S & Pad & Output \\
\hline
CN&4 conv &3*3&1&Yes&176*176*128\\
\hline
EG Block-1&  2 conv &3*3 &1&Yes&176*176*128\\
\hline
EG Block-2&  2 conv &3*3 &2&Yes&88*88*128\\
\hline
EG Block-3&  2 conv &3*3 &2&Yes&44*44*128\\
\hline
EG Block-4&  2 conv &3*3,5*5 &2,4&Yes&22*22*128\\
\hline
EG Block-5&  2 conv &5*5 &4&Yes&11*11*128\\
\hline
\end{tabular}}
\end{table}
\end{center}


\subsubsection{Contrast Features}
Saliency detection is related to the global or local contrast of foreground and background. 
For salient objects in an image, salient objects are the foreground that highlights the background around them. 
That is to say, salient features must be evenly distributed in the foreground, and the foreground and background regions are different. 
In order to capture such contrast information, we add contrast features related to each stage edge features $X^F_{i}$ outputted by the edge guidance block. 
Each contrast feature is computed by subtracting $X^F_{i}$ from its local average, where the size of local neighbor region is $3\times 3$. 
The average pooling operation that can reduce the errors caused by the variance of the estimated value due to the limited size of the neighborhood. Hence subtracting $AvgPool(X^F_{i})$ can retain more foreground information, and salient objects are easier to detect:
\begin{equation}
\label{E5}
X^c_{i}=X^F_{i}-AvgPool(X^F_{i}).
\end{equation}

\subsubsection{Deconvolution Features} 
After the last step, the sizes of the five contrast features is gradually decreasing. 
The first size of contrast feature $X^c_{1}$ is $176\times 176$, but the fifth size of contrast feature $X^c_{5}$ is $11\times 11$.
To obtain the same size of final output $176\times 176$, we need from back to front sequentially use five deconvolution (Deconv) layers to increase the sizes of  precomputed feature maps $X^F_{i}$ and $X^C_{i}$.  
At each Deconv blocks, we upsample the previous feature by the stride of 2. 
The result of the deconvolution feature map $D_{i}$ is computed by combining the information of edge feature $X^F_{i}$ and local contrast feature $X^C_{i}$.
We upsample the feature $D_{i}$ to obtain $D_{i+1}$. 
The Deconv operation is achieved by deconvolution layer with the kernel size of $5\times 5$ and the stride of 2. 
The input of this Deconv layer is the concatenation of $X^F_{i} , X^C_{i}$ and $D_{i+1}$ and the channels number of $D_{i+1}$ is sum of  $X^F_{i}$ and $D_{i+1}$,
\begin{equation}
\label{E6}
D_{i}=Deconv(X^F_{i}, X^C_{i}, D_{i+1}).
\end{equation}

\subsubsection{Local Features} 
The local block uses a convolution layer with a stride of 1 and a kernel size of $1\times 1$ to obtain the final local features maps $X^{L}$:
\begin{equation}
\label{E7}
X^{L}=Conv(X^F_{1}, X^C_{1}, D_{2}),
\end{equation} 
where the channel number of $X_{L}$ is the sum of sizes of $X^F_{1}$ and $D_{2}$, and the size of the local feature $X_{L}$ is $176\times 176$.
For convenience to fusion with the global feature, we need again use the Deconv operation to increase the size of the local feature $X^{L}$ from $176\times 176$ to $352\times 352$.

\subsubsection{Global Features} 
A good saliency detection model not only can capture local features but also can capture global features. 
Before assigning saliency tasks to a single small region, the saliency model needs to capture the global context of the image. 
To achieve this purpose, we use three convolution layers after the last VGG block to computer global feature $X_{G}$. 
The first two convolution layers use the kernel size of $5\times 5$, the last convolutional layer uses the kernel size of $3\times3$ and all convolutional layers have the same channel dimensions.

\subsection{Edge Guidance Block}
Five side outputs will generate five different scale features with different channels.
To effectively embed the edge feature $X^{E}_{i}$ into the feature $X_{i}$, we design a condition network to generate shared intermediate conditions in all layers for efficient computation, as shown in Fig.~\ref{2}. These generated conditions are regarded as one of inputs of the edge guidance block.

The inspiration of the edge guidance block comes from~\cite{Wang2018Recovering}, this paper employs deep spatial feature transform to recovery realistic texture in image super-resolution.
They use the possibility of semantic segmentation maps as the categorical prior, and a spatial feature transform layer is conditioned on semantic segmentation probability maps. 
Different from it, as shown in Fig.~\ref{2}, our edge guidance block is composed of two stage. 
In the first stage, our condition network uses four convolutional layers with the edge map as input to generate conditional features $X^{E}_{i}$, where each convolutional layer is with a kernel size of $3\times 3$ and a stride of 1.
The size of $X^{E}_{i}$ is $176\times 176$. 
Note that the condition network only produces coarse edge feature $X^{E}_{i}$. 
To better transferring edge features to next layers of the network and play a better guidance role, we propose to integrate the edge guidance block into hierarchical feature maps. 
The edge guidance block uses two separate branches to output two features $(\gamma, \beta)$ based on the conditional features. 
Then, we use $(\gamma, \beta)$ to transform $X_{i}$ into an edge-aware feature $X^{F}_{i}$ as follows:
\begin{equation}
\label{E4}
EGB(X^{F}_{i}|\gamma,\beta)=X_{i}\odot\gamma+\beta,
\end{equation}
where $\odot$ represents the element-wise product operation and $+$ represents the element-wise addition operation.
Since the spatial dimensions are preserved, the edge guidance block not only performs feature-wise manipulation but also spatial-wise transformation. 

\subsection{Loss Function}
Saliency detection and image segmentation usually come down to the optimization of non-convex energy functions consisting of data items and regularization items. 
A mathematical global model is the  Mumford-Shah (MS) model~\cite{Mumford1989Optimal}, which transforms the image segmentation problem into solving the minimum value of energy function. 
By constructing the energy function, the curve is evolved under the driving of the minimum value of energy function and the contour curve gradually approaches the boundary of the object, and the object is finally segmented. 
In~\cite{Luo2017Non}, they propose a supervised deep convolutional network, where the loss function approximates the MS functional are made of a cross entropy loss term between ground truth and predicted the saliency map and boundary loss term. 
The purpose of MS functional is to minimize IOU loss by maximizing the coincidence rate between predicted boundaries and real boundaries:
\begin{equation}
\label{E3}
F_{MS}\underbrace{\approx\sum_{j}\lambda_{j}\displaystyle\int_{v\epsilon\Omega_{j}}S_{j}(y(v),\widehat{y}(v))dv}_{cross~entropy~loss}+
\underbrace{\sum_{j}\gamma_{j}(1-IoU(C_{j},\widehat{C}_{j}))}_{boundary~loss}
\end{equation}
where $v$ is the pixel location, $S_{j}$ is the total loss between ground truth $(y)$ and predicted $(\widehat{y})$ saliency map, and IoU$(C_{j}, \widehat{C}_{j})$ is the intersection over union between the true boundary and predicted boundary. 
The positive constant $\lambda_{j}$ and $\gamma_{j}$ are used to tune the multi-criteria energy function in terms of data fidelity and total boundary length.

\begin{figure*}[!t]
\centering
\includegraphics[width=6.5in]{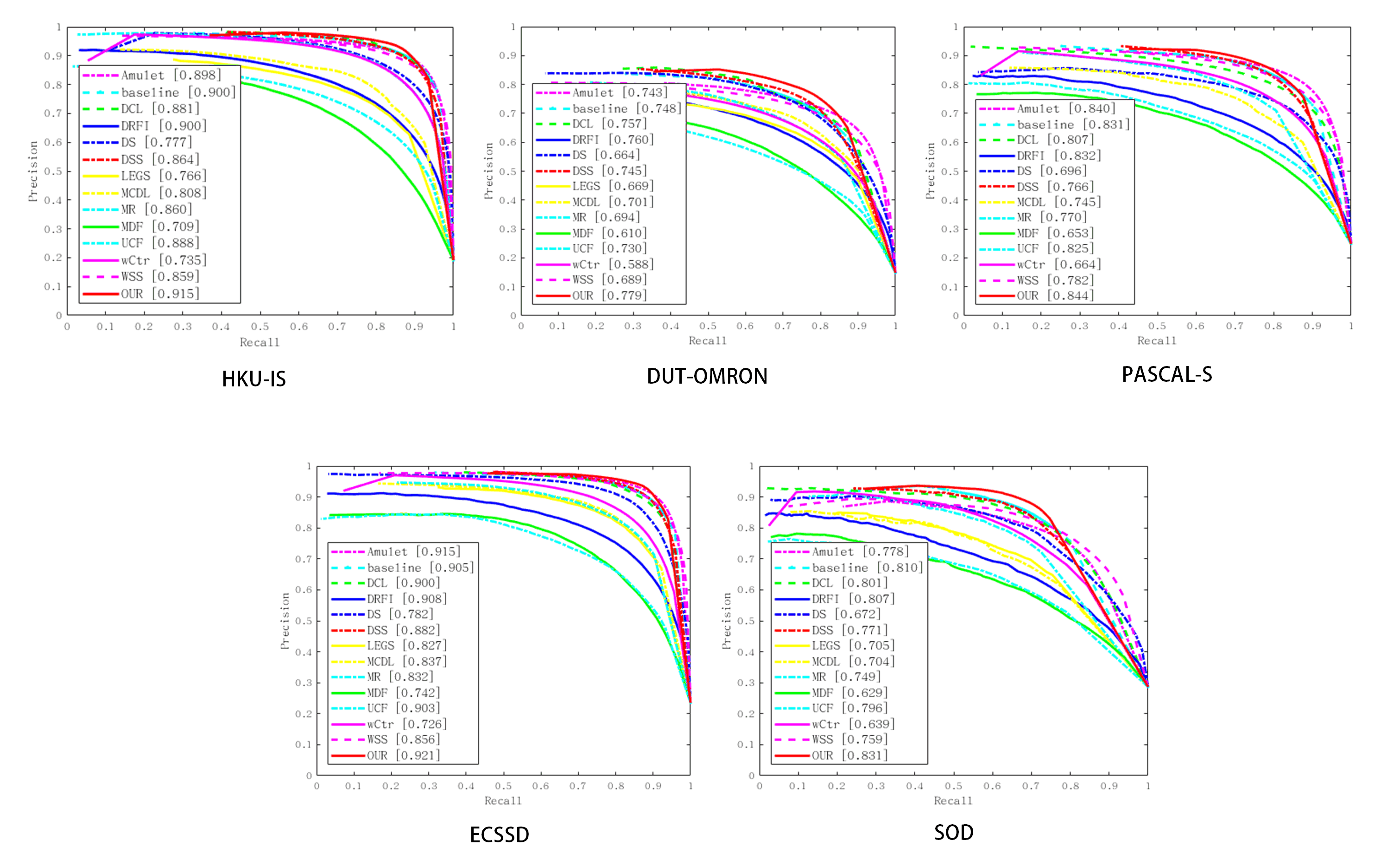}
\caption{Precision-recall curves for our model compared to MR~\cite{yang2013saliency}, DRFI~\cite{Wang2013Salient}, wCtr~\cite{zhu2014saliency}, MCDL~\cite{Rui2015Saliency}, MDF~\cite{li2015visual},DCL~\cite{Li2016Deep},DS~\cite{Li2016DeepSaliency}, DSS~\cite{Hou2016Deeply}, LEGS~\cite{Wang2015Deep}, UCF~\cite{Zhang2017Learning}, WSS~\cite{Wang2017Learning}, Amulet~\cite{Zhang2017Amulet}, NLDF~\cite{Luo2017Non}. Our model can deliver state-of-the-art performance on five datasets.}
\label{4}
\end{figure*}

\begin{center}
\begin{table*}
\caption{Maximum F-measure (larger is better) and MAE (smaller is better) of different saliency detection methods on five released saliency detection datasets.}\label{tab2}
\setlength{\tabcolsep}{4mm}{
\resizebox{\textwidth}{30mm}{
\begin{tabular}{|c|c|c|c|c|c|c|c|c|c|c|}
\hline
\multirow{2}{*}{*}  &\multicolumn{2}{|c|}{HKU-IS}     &\multicolumn{2}{|c|}{DUT-OMRON}            &\multicolumn{2}{|c|}{PASCAL-S}         &\multicolumn{2}{|c|}{ECSSD}  &\multicolumn{2}{|c|}{SOD}\\
 \hline
                                       & max $F_{\beta}$    & MAE            &max $F_{\beta}$      &MAE                  &max $F_{\beta}$    & MAE             & max $F_{\beta}$      &MAE             &max $F_{\beta}$    & MAE\\
\hline
ENFNet         &\textbf{0.915}                          &\textbf{0.040}         &\textbf{0.779}                          &\textbf{0.065}                &\textbf{0.843}                    &\textbf{0.095}             &\textbf{0.921}      &\textbf{0.051} &\textbf{0.831}                       &\textbf{0.134}\\
\hline
NLDF~\cite{Luo2017Non}  &0.902                           &0.048           &0.753                           &0.080               &0.831                      &0.099            &0.905                            &0.063            &0.810                        &0.143\\
\hline
Amulet~\cite{Zhang2017Amulet}     &0.899                            &0.050         &0.743                            &0.098                &0.839                     &0.099             &0.915                            &0.059            &0.778                        &0.156\\
\hline
WSS~\cite{Wang2017Learning}        &0.858                            &0.079           &0.689                          &0.110                  &0.782                   &0.141               &0.856                            &0.103            &0.759                      &0.181\\
\hline
UCF~\cite{Zhang2017Learning}         &0.888                          &0.061             &0.730                         &0.120                  &0.825                   &0.115                 &0.903                          &0.069              &0.796                      &0.159\\
\hline
LEGS~\cite{Wang2015Deep}       &0.770                             &0.118           &0.669                          &0.133                   &0.756                   &0.157                &0.827                          &0.118               &0.707                     &0.215\\
\hline
DSS~\cite{Hou2016Deeply}          &0.900                             &0.050           &0.760                         &0.074                    &0.832                    &0.104                &0.908                        &0.062                 &0.807                   &0.145\\
\hline
DS\cite{Li2016DeepSaliency}            &0.864                              &0.078           &0.745                          &0.120                    &0.766                   &0.176                 &0.882                     &0.122                     &0.771                 &0.199\\
\hline
DCL~\cite{Li2016Deep}         &0.892                                &0.054           &0.733                         &0.084                     &0.815                   &0.113                &0.887                       &0.072                  &0.795                   &0.142\\
\hline
MDF~\cite{li2015visual}       &0.861                                &0.076            &0.694                          &0.092                    &0.764                   &0.145                  &0.832                     &0.105                   &0.745                  &0.192\\
\hline
MCDL~\cite{Rui2015Saliency}     &0.808                               &0.092              &0.701                       &0.089                       &0.745                   &0.146                 &0.837                     &0.101                    &0.704                &0.194\\
\hline
wCtr~\cite{zhu2014saliency}        &0.735                               &0.138                &0.588                     &0.171                      &0.664                      &0.199                 &0.726                    &0.165                      &0.639             &0.231\\
\hline
DRFI~\cite{Wang2013Salient}      &0.777                                &0.144                 &0.664                     &0.150                     &0.696                         &0.210               &0.782                     &0.170                     &0.672            &0.242\\
\hline
MR~\cite{yang2013saliency}          &0.709                              &0.174                   &0.610                      &0.187                    &0.653                       &0.232                 &0.742                   &0.186                    &0.629            &0.274\\
\hline
\end{tabular}}}
\end{table*}
\end{center}

\begin{figure*}[!t]
\centering
\includegraphics[width=\textwidth]{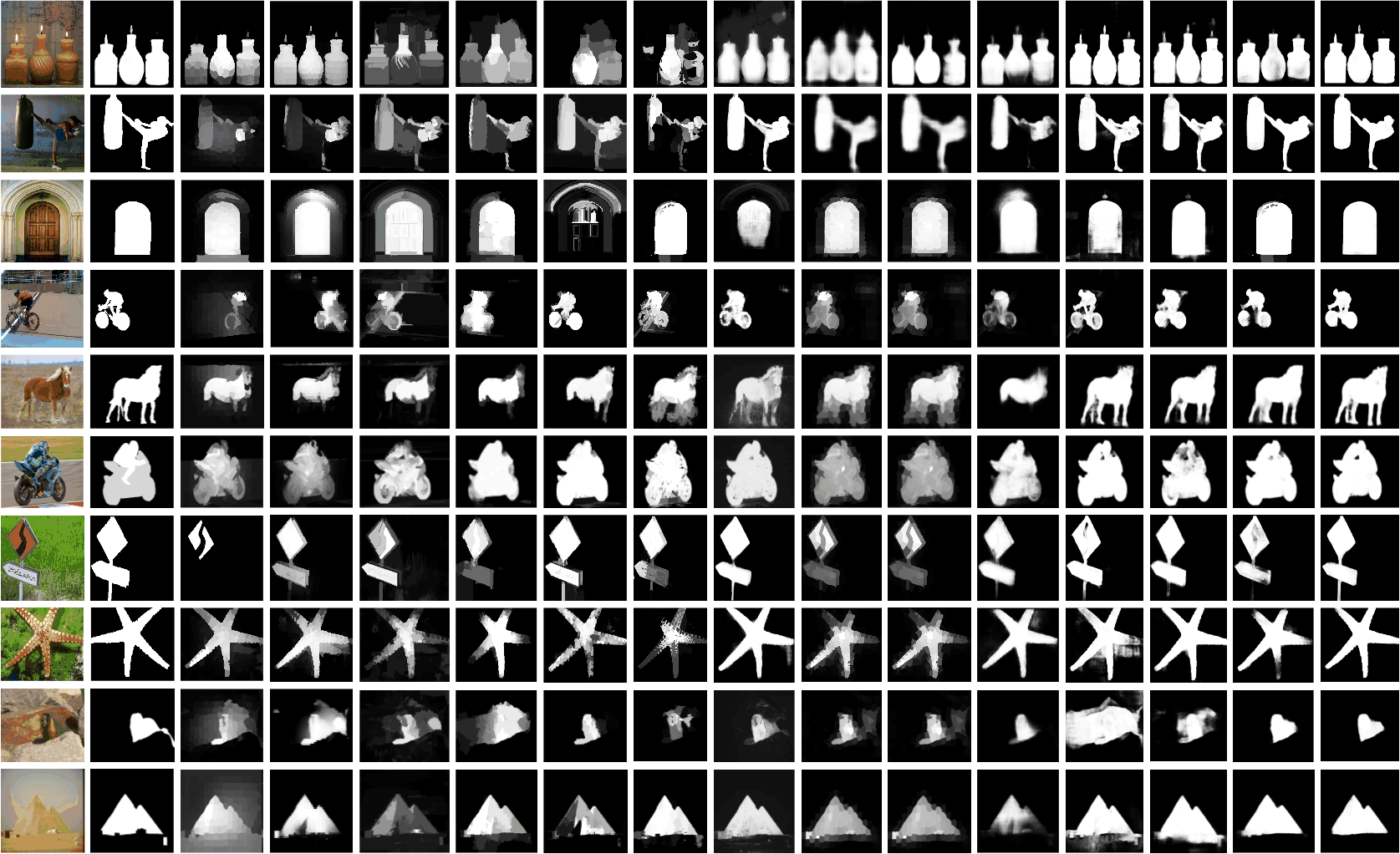}
\caption{Saliency maps produced by the MR~\cite{yang2013saliency}, DRFI~\cite{Wang2013Salient}, wCtr~\cite{zhu2014saliency}, MCDL~\cite{Rui2015Saliency}, MDF~\cite{li2015visual},DCL~\cite{Li2016Deep},DS\cite{Li2016DeepSaliency}, DSS~\cite{Hou2016Deeply}, LEGS~\cite{Wang2015Deep}, UCF~\cite{Zhang2017Learning}, WSS~\cite{Wang2017Learning}, Amulet~\cite{Zhang2017Amulet} and NLDF~\cite{Luo2017Non}. Our model can deliver state-of-the-art performance on five datasets. }
\label{3}
\end{figure*}

\subsubsection{Cross-entropy Loss} 
We use two linear operators $(W_{G},b_{G})$ and $(W_{L},b_{L})$ to combine the local and global features. 
And the softmax function is used to compute the probability for each pixel of  being salient or not:
\begin{equation}
\label{E8}
\hat{y}=p(y(v)=s)=\frac{e^{W^s_{L}X_{L(v)}+b^s_{L}+W^s_{G}X_{G}+b^s_{G}}}{\sum_{s\prime\in\{0,1\}}e^{W^{s\prime}_{L}X_{L(v)}+b^{s\prime}_{L}+W^{s\prime}_{G}X_{G}+b^{s\prime}_{G}}}
\end{equation}
where $p$ is the probability for each pixel of being salient, $v$ is the location of a pixel and $s$ indicates that the pixel belongs to the foreground.
The cross-entropy loss function is:
\begin{equation}
\label{E9}
G_{j}(y(v),\hat{y}(v))=-\frac{1}{N}\sum^{N}_{i=1}\sum_{s\in\{0,1\}}{(y(v_{i})=s)(\log(\hat{y}(v_{i}=s))}
\end{equation}
where $y(v)$ is the ground truth and $\hat{y}(v)$ is the predicted saliency map.

\subsubsection{IoU Boundary Loss} 
This inspiration comes from the applications of the IOU boundary loss in image segmentation~\cite{Milletari2016V, Taha2015Metrics}. 
IOU boundary loss computes the error between the true boundary $C_{j}$ and predicted boundary $\hat{C}_{j}$.
The boundary pixels use a Sobel operator followed by a tanh activation function and this active function predicts the gradient magnitude of saliency maps to a probability range of [0,1]. 
The IOU boundary loss can be represented as:
\begin{equation}
\label{E10}
IoU Loss=1-\frac{2\mid{C_{j}\cap{\hat{C}_{j}}}\mid}{\mid{C_{j}}\mid+\mid{\hat{C}_{j}}\mid}
\end{equation}

\begin{figure*}[!htbp]
\centering
\includegraphics[width=6.5in]{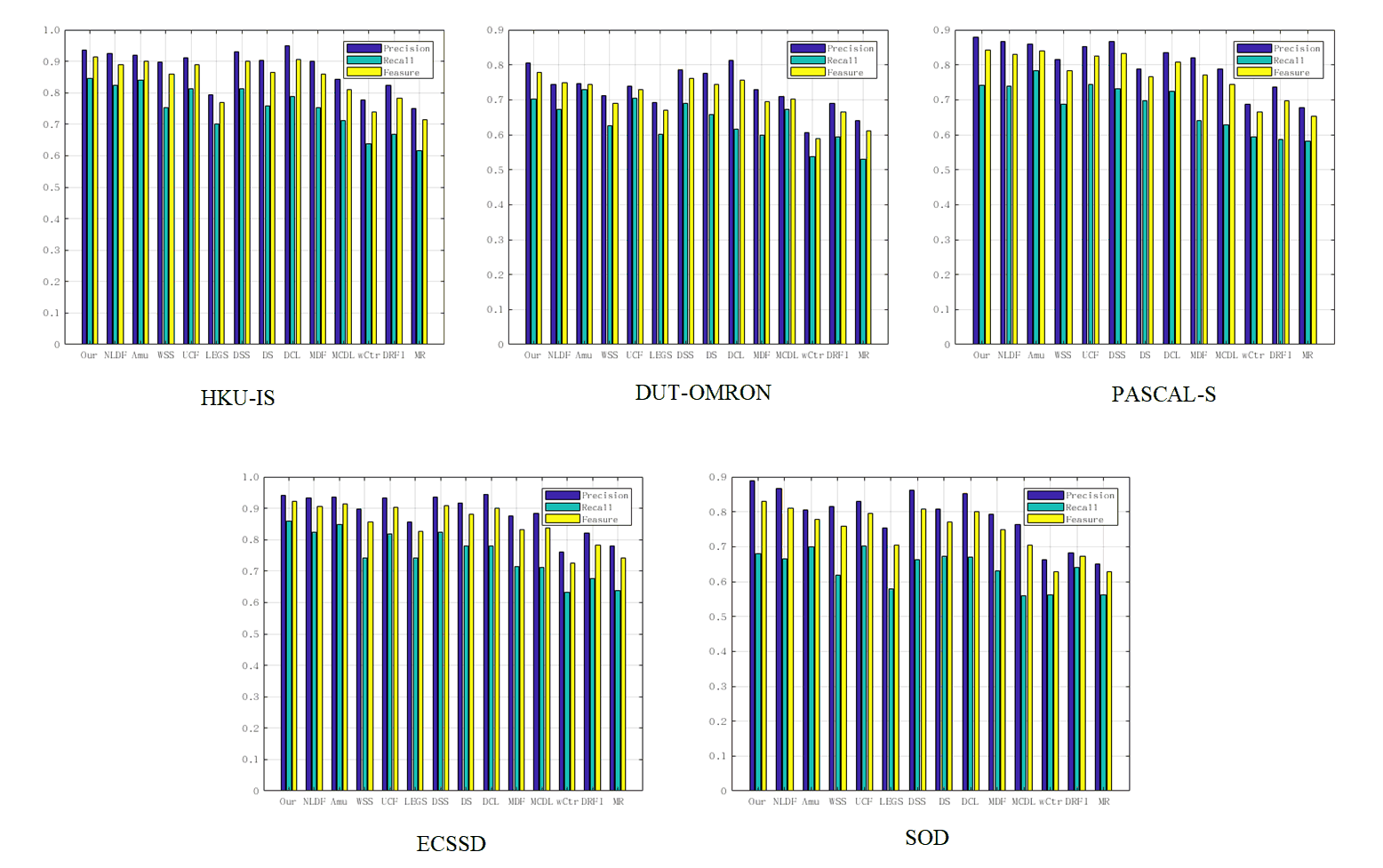}
\caption{For a better visual effect, we represent the Precision, Recall and F-measure value in the form of  a histogram compared to MR~\cite{yang2013saliency}, DRFI~\cite{Wang2013Salient}, wCtr~\cite{zhu2014saliency}, MCDL~\cite{Rui2015Saliency}, MDF~\cite{li2015visual}, DCL~\cite{Li2016Deep}, DS~\cite{Li2016DeepSaliency}, DSS~\cite{Hou2016Deeply}, LEGS~\cite{Wang2015Deep}, UCF~\cite{Zhang2017Learning}, WSS~\cite{Wang2017Learning}, Amulet~\cite{Zhang2017Amulet},  NLDF~\cite{Luo2017Non}. Our model is in the first column and delivers state-of-the-art performance on five datasets.}
\label{5}
\end{figure*}

\subsection{Implementation}
The pre-trained VGG-16 model~\cite{simonyan2014very} is used to initialize the weights in first five VGG Blocks. 
All weights of convolution and deconvolution are initialized randomly with a constant (0.01), and the biases are initialized to 0. 
The learning rate is $10^-5$ and $\lambda_{j}$,  $\gamma_{j}$ in~\eqref{E3} were set to 1.

We train our model on the MSRA-B~\cite{liu2010learning} dataset and test it on five popular benchmark datasets. 
The same as the NLDF~\cite{Luo2017Non}, we use horizontal flipping as data augmentation, resulting in an augmented image set with twice large than the original one. 
All training images are resized to $352\times 352$ and use the cross entropy loss and the IOU boundary loss to optimize our network. 
Without further optimization, the trained model is used to predict the saliency maps in other datasets. 
The whole training produces for 10 epochs with a single image batch size, and it takes about ten hours to finish the whole training.

\section{Experiments}
We have implemented our network on a single Nvidia GTX 1070Ti GPU and in Tensorflow~\cite{abadi2015tensorflow}, and our method maintains a fast runtime of 0.08 second per image.
This section will present the details of our evaluation settings, comparison with the state-of-the-art methods and impact of the proposed edge guidance block on the detection performance.

\subsection{Experimental Setup}

{\flushleft \bf Evaluation Datasets.}
We evaluate the proposed model on five public benchmark datasets including
HKU-IS~\cite{li2015visual}, PASCAL-S~\cite{li2014secrets}, DUT-OMRON~\cite{yang2013saliency}, ECSSD~\cite{yan2013hierarchical} and SOD~\cite{martin2001database}. Their details are as follows.

\begin{itemize}
\item \emph{HKU-IS:} This dataset contains 4447 images with high quality pixel labeling, and most of the images have low contrast and many salient objects. 
It also contains many independent salient objects or objects touching the image boundary.

\item \emph{PASCAL-S:} This dataset has 850 natural images which are generated from the PASCAL VOC~\cite{Everingham2006The} segmentation challenge. 
The ground truths labeled by 12 experts contains both pixel-wise saliency and eye fixation.

\item \emph{DUT-OMRON:} This is a large dataset with 5168 high quality images. 
Each image in this dataset has one or more salient objects and a cluttered background. 
Therefore, this dataset is more difficult and challenging, which provides more space for improvement for the research of saliency detection.

\item \emph{ECSSD:} This dataset contains 1000 natural and complex images with pixel-wise ground truth annotations and these images are manually selected from the Internet.

\item \emph{SOD:} This dataset has 300 images, and it was originally designed for image segmentation. 
It is challenging because many images have multiple objects which with low contrast or touching the image boundary.

\end{itemize}

{\flushleft \bf Evaluation Criteria.}
We use three evaluation metrics to evaluate the performance of our model with other salient object detection methods, including 
Precision-recall(PR) curves, F-measure score and mean absolute error(MAE)~\cite{Borji2012Salient}. 
With continuous values normalized the saliency map to the range of 0 to 255, we compute the corresponding binary map, then compute the precision /recall pairs of all boundary maps in the dataset. 
The PR curve of a dataset explains the mean precision and recall of saliency maps at different thresholds. 
The F-measure is a harmonic mean of average precision and average recall, F-measure can be represented as:
\begin{equation}
\label{E11}
F_{\beta}=\frac{(1+\beta^2)\cdot{Precision}\cdot{Recall}}{\beta^2\cdot{Precision}\cdot{Recall}}
\end{equation}
where $\beta^2$=0.3 to emphasize precision over recall the same as~\cite{Achantay2009Frequency}. 
The mean absolute error (MAE) is to measure the average difference between predicted saliency map S and ground truth G:
\begin{equation}
\label{E12}
MAE=\frac{1}{W\times{H}}\sum^W_{x=1}\sum^H_{y=1}\mid{S(x,y)}-G(x,y)\mid
\end{equation}
where W and H is the width and height of a given image.

\begin{figure}[htbp]
\centering
\includegraphics[width=3.2in]{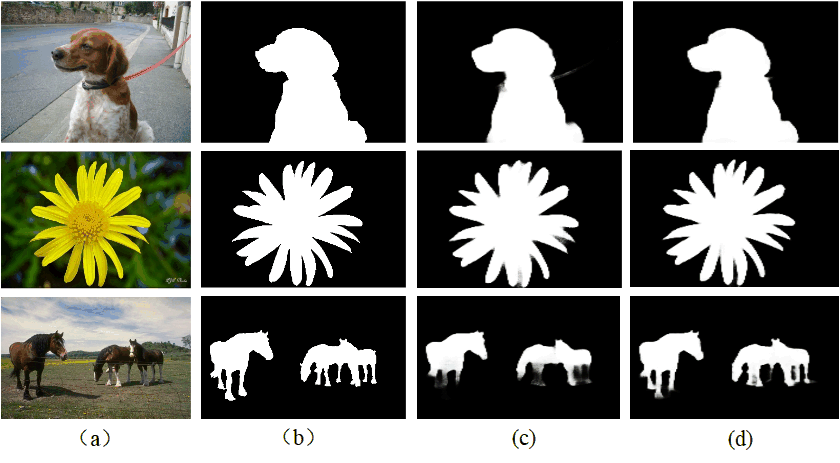}
\caption{Three examples of effectiveness validation of edge information. (a) Input images. (b) Ground truths. (c) Saliency results without edge guidance. (d) Saliency results with edge guidance. }
\label{6}
\end{figure}

\subsection{Comparison with State-of-the-art Methods}
To verify the effectiveness of our experiments, we compare our method with other state-of-art ones including three conventional methods (MR~\cite{yang2013saliency}, DRFI~\cite{Wang2013Salient}, wCtr~\cite{zhu2014saliency}) and en deep learning based methods (MCDL~\cite{Rui2015Saliency}, MDF~\cite{li2015visual}, DCL~\cite{Li2016Deep}, DS~\cite{Li2016DeepSaliency}, DSS~\cite{Hou2016Deeply}, LEGS~\cite{Wang2015Deep}, UCF~\cite{Zhang2017Learning}, WSS~\cite{Wang2017Learning}, Amulet~\cite{Zhang2017Amulet}, NLDF~\cite{Luo2017Non}). It is worth mentioning that the method Amulet~\cite{Zhang2017Amulet} is also utilize edge information. Note that since some of the HKU-IS datasets are used for training in MDF~\cite{li2015visual}, we only calcute the evaluation metrics on the test dataset on HKU-IS. 
Because MDF~\cite{li2015visual} also provides 200 saliency maps on the SOD dataset. Therefore, the evaluation of other methods on SOD uses only 200 saliency maps, originally 300 images.
For fair comparison, we use either the implementations with suggested parameter settings and the saliency maps provided by the authors.

\begin{center}
\begin{table}[htbp]
\caption{Effects of different number of the edge guidance blocks (EGB) on the performance.}\label{tab3}
\setlength{\tabcolsep}{1.6mm}{
\begin{tabular}{|c|c|c|c|c|c|c|}
\hline
*                                      &\multicolumn{2}{|c|}{Five EGB}     &\multicolumn{2}{|c|}{Three EGB}            &\multicolumn{2}{|c|}{Zero EGB}         \\
\hline
                                       & max $F_{\beta}$    & MAE            &max $F_{\beta}$      &MAE                  &max $F_{\beta}$    & MAE            \\
\hline
HKU-IS         &\textbf{0.915}                          &\textbf{0.040}         &0.906                          &0.046               &0.902                  &0.048            \\
\hline
DUT-OMRON  &\textbf{0.779}                           &\textbf{0.065}           &0.772                           &0.077               &0.753                      &0.080            \\
\hline
PASCAL-S  &\textbf{0.843}                           &\textbf{0.095}           &0.839                           &0.098               &0.831                      &0.099            \\
\hline
ECSSD     &\textbf{0.921}                            &\textbf{0.051}          &0.913                            &0.057              &0.905                      &0.063            \\
\hline
SOD       &\textbf{0.831}                            &\textbf{0.134}          &0.828                            &0.136               &0.810                     &0.143             \\
\hline
\end{tabular}}
\end{table}
\end{center}

We compare the proposed methods with the others in terms of F-measure scores, MAE scores, and PR-curves. 
The results of quantitative comparison with compared methods are reported in Fig.~\ref{4} and Table~\ref{tab2}. 
For the PR curves, we also achieve best performance compared to other methods on five datasets, as shown in Fig.~\ref{4}. 
In particular, for the top-level method DCL, which uses CRF-based post-processing steps to refine the resolution, our method still achieves the better performance. 
To verify the effectiveness of our ENFNet, we further report the results of F-measures and MAE scores in Table~\ref{tab2}.
From the results we can clearly see that our approach is clearly superior to competitive methods in terms of F-measures and MAE scores, especially in challenging datasets. 
In terms of F-measures, our ENFNet increases the value of F-measure by 2.6\% and 2.1\% over the second best method NLDF on the two most challenging datasets DUT-OMRON and SOD respectively. 
For MAE values, our method decreases 15\% and 9\% over the second best method NLDF on the two most challenging datasets DUT-OMRON and SOD respectively. 
Experimental results show that the edge guidance is effective.

To express the superiority of our method more intuitively, we report the values of Precision, Recall and F-measure in the form of the histogram with five datasets and 13 methods, as shown in Fig.\ref{5}.
On the five datasets, our method can achieve best performance compared to 13 methods.

For a better visual experience, Fig.~\ref{3} provides a visual comparison of our approach and other methods. 
It can be seen our method generates more clear saliency maps in various challenging images.
We select two most representative images from each dataset to display the saliency map. 
These images usually contain multiple significant targets and have complex backgrounds or unclear edges. 
Thus such images are more responsive to the superiority of our approach. 
From Fig.~\ref{3} we can see that our method obtains the best results which are much close to the ground truth in various challenging scenarios. 
To be specific, with the help of edge information guidance, the proposed method not only highlight the salient object regions clearly, but also generate the saliency maps with clear boundaries and consistent saliency values.

\subsection{Impact of Edge Guidance Block}
Because we use the edge information in our method, we compare to an edge-related method Amulet~\cite{Zhang2017Amulet} which achieves good performance. 
To verify the effectiveness of our proposed edge guidance block, the inclusion of edge feature in our model as compared to baseline~\cite{Luo2017Non} for an increase in max $F_{\beta}$  and decrease in MAE on HKU-IS, DUT-OMRON, PASCAL-S, ECSSD and SOD five datasets as shown in Table~\ref{tab2}.
It can be observed visualization results that without edge guidance block in the NLDF~\cite{Luo2017Non} are shown in Fig.\ref{3}. 
With the help of the edge guidance block, our approach can better localize the boundaries of most salient objects and produce more precise saliency maps. 
Several visual examples are showed in Fig.~\ref{6}. 

In addition, we also conduct an experiments to show the impact of the number of edge guidance blocks on the saliency detection performance. 
One is that only three edge guidance blocks are used in the first three layers, and another is to remove all edge guidance blocks.
The results are shown in Table~\ref{tab3}.
From the results we can see that our complete method ENFNet achieves best than other variants, demonstrating the effectiveness of our design on the network.

\section{Conclusion}
In this paper, we have proposed a novel Edge-guided Non-local FCN for salient object detection from the local and global perspective. 
We demonstrated that with saliency edge prior knowledge to perform edge-guided feature learning is beneficial to generating high-quality saliency values. 
Through the edge guidance block the edge features are embedded into feature learning of our network, our method can preserve more accurate edge structure information. 
Experimental results demonstrated that our proposed method consistently improves the performance on all five benchmarks and outperforms 13 state-of-the-art methods under different evaluation metrics.
In future work, we will study other prior knowledge and information like semantic priors~\cite{li19ec-cnn} and thermal infrared data~\cite{li2019rgb} to help improving the performance of salient object detection, and also extend our framework to video object saliency detection.




\ifCLASSOPTIONcaptionsoff
  \newpage
\fi

\bibliographystyle{IEEEtran}
\bibliography{mybib}

\end{document}